\documentclass[runningheads]{llncs}
\usepackage{accv}
\usepackage[accsupp]{axessibility}
\usepackage{accvabbrv}
\usepackage{amsmath}
\usepackage{booktabs}
\usepackage{dcolumn}
\usepackage{graphicx}
\usepackage{multirow}
\usepackage{pifont}
\usepackage{tikz}
\usepackage{hyperref}
\usepackage{orcidlink}
\usepackage{makecell}

\newcommand{\cmark}{\ding{51}}
\newcommand{\xmark}{\textcolor{red}{\ding{55}}}

\DeclareMathOperator*{\argmin}{arg\,min}

\makeatletter
\renewcommand{\paragraph}{%
  \@startsection{paragraph}{4}%
  {\z@}{-0.5em}{-0.5em}%
  {\normalfont\normalsize\bfseries}%
}
\makeatother

\begin{document}
\title{3D-Aware Instance Segmentation and Tracking \\ in Egocentric Videos}

\titlerunning{3D-Aware Instance Segmentation and Tracking in Egocentric Videos}
\author{Yash Bhalgat\orcidlink{0000-0001-7775-6250}\thanks{Equal contribution.}$^{1}$ \!\quad
Vadim Tschernezki\protect\footnotemark[1]$^{1,2}$ \!\quad
Iro Laina$^{1}$ \!\quad Jo\~{a}o F.\ Henriques$^{1}$ \\ Andrea Vedaldi$^{1}$ \quad Andrew Zisserman$^{1}$}

\authorrunning{Y.~Bhalgat \& V.~Tschernezki et al.}
\institute{Visual Geometry Group, University of Oxford \and NAVER LABS Europe \\
\email{\{yashsb,vadim,iro,joao,vedaldi,az\}@robots.ox.ac.uk}}

\maketitle
\begin{abstract}
  Egocentric videos present unique challenges for 3D scene understanding due to rapid camera motion, frequent object occlusions, and limited object visibility.
  This paper introduces a novel approach to instance segmentation and tracking in first-person video that leverages 3D awareness to overcome these obstacles. Our method integrates scene geometry,
  3D object centroid tracking, and instance segmentation to create a robust framework for analyzing dynamic egocentric scenes.
  By incorporating spatial and temporal cues, we achieve superior performance compared to state-of-the-art 2D approaches.
  Extensive evaluations on the challenging EPIC Fields dataset demonstrate significant improvements across a range of tracking and segmentation consistency metrics.
  Specifically, our method outperforms the next best performing approach by $7$ points in Association Accuracy (AssA) and $4.5$ points in IDF1 score, while reducing the number of ID switches by $73\%$ to $80\%$ across various object categories.
  Leveraging our tracked instance segmentations, we showcase downstream applications in 3D object reconstruction and amodal video object segmentation in these egocentric settings.
\end{abstract}

\newcommand{\todo}[1]{\textcolor{orange}{[\textbf{TODO\@:} #1]}}

\section{Introduction}%
\label{sec:intro}

Egocentric videos, which capture the world from a first-person perspective, are a focus of increasing attention in computer vision due to their importance in  applications such as augmented reality and robotics.
Among various tools for video analysis, object tracking is of particular importance, but also faces significant challenges, in the egocentric case.
Most video object segmentation (VOS) methods~\cite{cheng2023tracking,masa,meinhardt2022trackformer,wu2022seqformer}, in fact, assume that the videos contain slow, steady camera motions that keep the view centered on the object of interest~\cite{davis2019,davis2016,athar2023burst}.
In comparison, egocentric videos are taken from a first-person perspective, where the camera wearer's movements introduce rapid and unpredictable changes in viewpoint.
Additionally, objects frequently move in and out of the field of view, and thus are often partially or wholly occluded and/or truncated.

For example, in the EPIC KITCHENS dataset~\cite{Damen2021PAMI}, the person recording the video might move a \textit{pan} on top of a hob and leave it there for several minutes while moving around in the kitchen.
During that time, they might observe more objects that look similar to the pan, which may cause an algorithm to incorrectly associate them to the pan itself.
In general, video segmenters tend to lose track of the object partially or entirely due to occlusion or truncation.
These issues are exacerbated when tracking multiple objects simultaneously.

Existing state-of-the-art video object segmenters try to overcome these limitations by aligning segments with dense or sparse correspondences.
These are obtained from optical flow or point tracking~\cite{rajivc2023segment} and serve as a proxy for spatial reasoning.
However, these methods can establish correspondences only in relatively short video windows due to their computational cost and poor reliability during severe viewpoint changes.
The result are fragmented and incomplete object tracks, which limit their usefulness, particularly in egocentric videos.

In order to address these shortcomings, we can look at how humans locate objects.
An important cue that helps correct reassociation is \emph{object permanence}, a concept that human infants develop very early~\cite{santrock2002topical}.
Permanence captures the idea that objects do not cease to exist when they are not visible.
Combined with spatial awareness, this means that the 3D location of objects at rest should not change when they are out of view or occluded. It has previously been explored for egocentric videos in `Out of Sight, Not Out of Mind' (OSNOM)~\cite{plizzari2023}.

This brings us to the question of how to incorporate such spatial awareness in an object tracking algorithm.
We achieve this by extracting scene geometry from the video stream and using it as an additional supervisory signal to refine tracks produced by a video segmentation model.
More specifically, we obtain depth maps and camera parameters for the frames of the video and use this information to calculate the 3D location of the object instances.
We then propose a novel approach for refining instance segmentation and tracking in egocentric videos that leverages 3D awareness to overcome the limitations of 2D trackers.
By integrating a scene-level 3D reconstruction, coarse 3D point tracking, and 2D segmentation, we obtain a robust framework for analyzing dynamic egocentric videos.
In particular, by incorporating both spatial \textit{and} temporal cues from the 3D scene, our method handles occlusions and re-identifies objects that have been out of sight for some time, leading to more consistent and longer object tracks.

Our experiments on the challenging EPIC Fields dataset~\cite{EPICFields2023} demonstrate significant improvements in tracking accuracy and segmentation consistency compared to state-of-the-art video object segmentation approaches.
Furthermore, we showcase the potential of our method in downstream applications such as 3D object reconstruction and amodal video object segmentation, where the consistent and accurate object tracks produced by our method enable more accurate and complete reconstructions.

\section{Related Work}%
\label{sec:relwork}

\subsubsection{Video object segmentation.}

Video object segmentation (VOS) has seen significant advancements over the past decade~\cite{zhou2022survey}, driven by the need to accurately segment and track objects across video frames.
Traditional methods often relied on frame-by-frame processing, which struggled with maintaining consistent object identities over long sequences.
Early approaches such as MaskTrack R-CNN~\cite{yang2019video} and FEELVOS~\cite{voigtlaender2019feelvos} introduced the concept of using temporal information to improve segmentation consistency.
MaskTrack R-CNN extended Mask R-CNN to video by adding a tracking head that links instances across frames, while FEELVOS utilized a pixel-wise matching mechanism to propagate segmentation masks.
The introduction of memory networks and attention mechanisms marked a significant leap in performance.
STM~\cite{oh2019video}, AOT~\cite{yang2021associating} and XMem~\cite{cheng2022xmem} leveraged memory networks to store and retrieve information across frames, enabling more robust handling of occlusions and reappearances.
Many recent works~\cite{choudhuri2023context,choudhuri2021assignment,qiao2021vip,wang2021end} have proposed end-to-end approaches for video object segmentation as well as panoptic segmentation.
VisTR~\cite{wang2021end}  and SeqFormer~\cite{wu2022seqformer} employed transformers to model long-range dependencies and global context.
VisTR treated video segmentation as a direct set prediction problem, while SeqFormer introduced a sequential transformer architecture that processes video frames in a temporally coherent manner.

Additionally, methods like DEVA~\cite{cheng2023tracking} employed decoupled video segmentation approaches, combining image-level segmentation with bi-directional temporal propagation to handle diverse and data-scarce environments effectively.
This also helps tackle open-vocabulary settings.
MASA~\cite{masa} uses the Segment Anything Model (SAM) as a robust segment proposer, and learns to match segments that correspond to the same object.
An adapter can be trained to map those segments to a closed set of classes, in zero-shot settings.

\subsubsection{Point tracking-based methods.}

Point tracking-based methods have been pivotal in advancing VOS by providing a means to establish correspondences across frames.
Many powerful point trackers have been recently proposed such as TAP-Vid~\cite{doersch2022tap} benchmark that focused on tracking physical points in a video and works such as CoTracker~\cite{karaev2023cotracker} and PIP~\cite{harley2022particle}.
CenterTrack~\cite{zhou2020tracking} combined object detection with point tracking, leveraging the strengths of both approaches.
TAPIR~\cite{doersch2023tapir} trains an initial matching network (analogous to SeqFormer) and an iterative refinement network (which focuses on continuous adjustments to predicted points' positions), using synthetic data, to predict accurate point tracks.
SAM-PT~\cite{rajivc2023segment} is a point-centric interactive video segmentation method, which propagates a sparse set of points, chosen by a user, to other frames.

\subsubsection{3D-informed instance segmentation and tracking.}

A recent line of work closely related to the problem we address here involves lifting and fusing inconsistent 2D labels or segments into 3D models.
In particular, Panoptic Lifting~\cite{siddiqui2023panoptic}, ContrastiveLift~\cite{bhalgat2024contrastive}, PVLFF~\cite{chen2024panoptic}, and Gaussian Grouping~\cite{gaussian_grouping} employ mechanisms for 3D instance segmentation in \emph{static} scenes.

Operating under the assumption that objects remain stationary, they show that a 3D reconstruction of the scene enables the fusion of unassociated 2D instances (\ie, inconsistent instance identities across views) using Hungarian matching~\cite{siddiqui2023panoptic}, contrastive learning~\cite{bhalgat2024contrastive,chen2024panoptic} or video object tracking~\cite{gaussian_grouping,cheng2023tracking}.
Instead of instance segmentation and tracking, GARField~\cite{kim2024garfield}, OmniSeg3D~\cite{ying2024omniseg3d}, and N2F2~\cite{bhalgat2024n2f2} focus on 3D hierarchical grouping, a problem which also requires resolving ambiguities that arise when fusing conflicting multi-view masks (such as those obtained by the Segment Anything Model~\cite{kirillov2023segment}).

Exploiting 3D information in egocentric videos has been less explored due to the challenges of reconstructing dynamic objects.
Following~\cite{bhalgat2024contrastive}, EgoLifter~\cite{gu2024egolifter} uses contrastive learning to lift 2D segmentations to 3D, while also using a transient prediction network to handle dynamic objects.
Plizzari~\etal~\cite{plizzari2023} focus specifically on 3D tracking of {\em dynamic} objects, rather than segmenting or reconstructing them. They form 3D centroid tracks by lifting 2D centroids to 3D and matching observations based on 3D distance and visual similarity.
We follow~\cite{plizzari2023}, in that we lift objects to 3D using estimated depth, and initialise, match and update tracks based on 3D location and DINOv2~\cite{oquab2023dinov2} feature similarity. However, 
we also incorporate instance and category information from a base VOS model into our cost formulation, creating a more robust 3D-aware object tracking system that excels in refining imperfect or noisy input 2D object tracks, achieving superior long-term object consistency as compared to existing 2D tracking methods.

\section{Method}%
\label{sec:method}

\DeclareRobustCommand{\eye}{%
  \begingroup\normalfont
  \includegraphics[height=\fontcharht\font`\B]{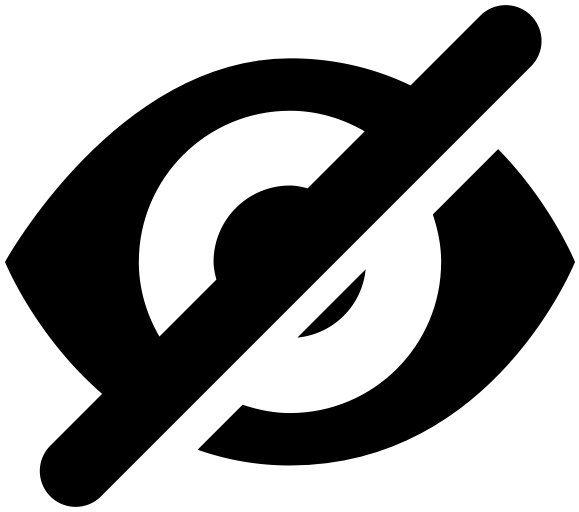}%
  \endgroup
}
\begin{figure}[t]
    \centering
    \includegraphics[width=\textwidth]{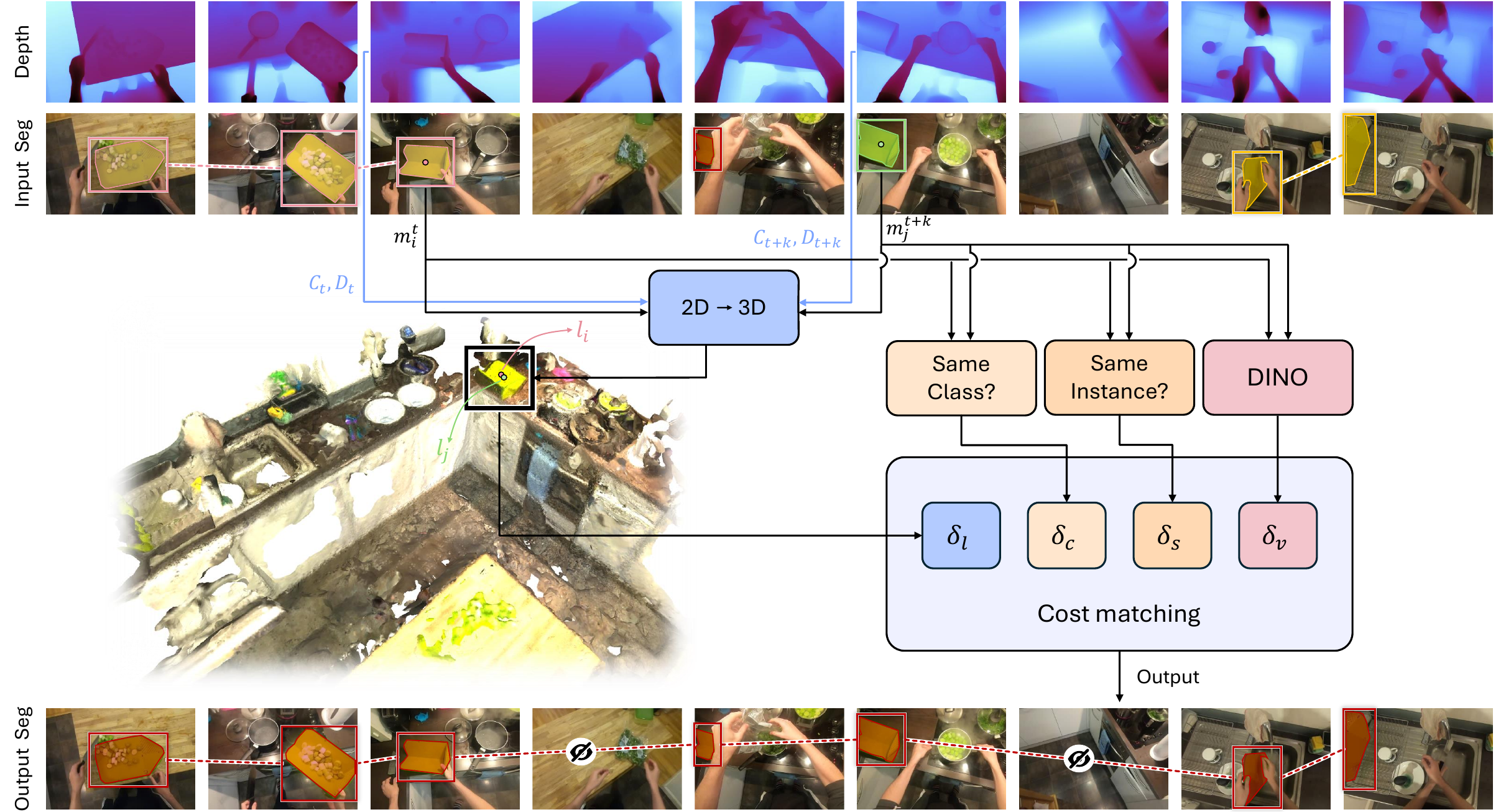}
    \caption{Overview of the proposed method for 3D-aware object tracking in egocentric videos.
    The method begins by taking image-level segments and object tracks from a pre-trained video object segmentation model, which are then lifted to 3D using per-frame depth estimates and scene geometry.
    These segments are fused across time with a 3D-aware tracking cost formulation to refine and maintain consistent object identities throughout the video sequence, even when the objects go out of sight (indicated by \eye{}).}%
    \label{fig:method}
\end{figure}

Given an egocentric video, our objective is to obtain long-term consistent object tracks by leveraging 3D information as well as an initial set of object segments and tracks obtained from a 2D-only video object segmentation (VOS) model.
Our proposed method overcomes the limitations of 2D VOS models in maintaining \emph{long-term consistent} object identities in egocentric scenarios and produces object tracks that persist despite severe occlusion and objects intermittently moving out of sight.

\Cref{fig:method} provides a high-level overview of the method. We take as input an initial set of image-level segments and object tracks obtained from a pretrained VOS model.
Then, we lift these 2D segments into 3D using per-frame depth from a pretrained depth estimator along with scene geometry information, and link them across time using our proposed tracking cost formulation.
We first define the above problem statement more concretely in \Cref{sec:prob-statement}, and the 3D-aware tracking algorithm in \Cref{sec:algorithm}.
Then, we describe our design that includes different attributes we extract for the 2D segments in \Cref{sec:attributes}, followed by our cost formulation in \Cref{sec:cost-matching}.

\subsection{Problem statement}%
\label{sec:prob-statement}

We begin with an egocentric video sequence consisting of $N$ frames $\mathbf{I}^t$, $t \in \{1, \dots, N\}$, along with the output of an off-the-shelf 2D VOS model. The objective of the method is to compute a set of tracks for the entire video
$\{\mathbf{T}^N_i\}$ with associated segment IDs $\{\tilde{s}_i^N\}$ that have the desired temporal consistency.
The initial output contains a set of object tracks that, while partially correct, often contain errors --- particularly when objects temporarily leave the field of view or are occluded.
Our goal is to \emph{refine} and \emph{reassemble} these tracks, leveraging 3D information to correct errors and achieve more consistent long-term tracking.
Crucially, we don't discard the initial track IDs obtained from the 2D-only VOS model.
Instead, we incorporate this information into our refinement process, using it as a valuable prior for maintaining object identities.
In this manner, we go beyond the previous 3D aware matching, initialisation and matching method~\cite{plizzari2023} that we build upon.

\subsection{3D aware tracking}%
\label{sec:algorithm}

First, we decompose the initial tracks into per-frame segments $\mathbf{M}^t=\{ m_i^t \mid 1 \leq i \leq |\mathbf{M}^t| \}$.
Specifically, each $\mathbf{M}^t$ contains a set of 2D segments $m^t_i$, representing the objects detected in frame $t$.
For each segment $m^t_i$, we compute an attribute vector $\mathbf{b}^t_i = (b_{i,1}^t, b_{i,2}^t, \ldots, b_{i,n}^t)$ that encodes various characteristics of the segment including its initial ID $s_i^t$ from the 2D VOS model, 3D location, visual features, and category information.
These attribute vectors play a crucial role in our method, as they allow us to establish correspondences between segments across frames.

We employ a frame-by-frame track refinement approach using the Hungarian algorithm.
At each frame $t$, we consider the existing tracks $\mathbf{T}^{t-1}$ formed in the previous $t-1$ frames and new segments $\mathbf{M}^t$ from the current frame $t$.
The $i$-th track within $\mathbf{T}^{t-1}$ is associated with an attribute vector $\widetilde{\mathbf{b}}_i^{t-1}$, computed as an aggregate of the attributes of segments assigned to it (\textit{c.f.}~\cref{sec:track-attr}), and \emph{refined} segment ID $\tilde{s}_i^{t-1}$.
We match the new segments at time $t$ to the tracks $\mathbf{T}^{t-1}$ by solving the following optimization problem to obtain the \emph{new refined} segment IDs $\{s^t_i\}$:
\begin{equation}
\argmin_{\{s^t_i\}} \sum_{i,j} J(s^t_i, \tilde{s}^{t-1}_j, \mathbf{b}^t_i, \widetilde{\mathbf{b}}^{t-1}_j)\\
\end{equation}
subject to $s^t_i \in \{1, \dots, S\}$ and $s^t_i\neq s^t_j$ if $i\neq j$, where $S$ is the total number of unique object identifiers.
The second condition enforces that no two segments in the same frame can have the same identifier. 
The cost function $J$ is defined as:
\begin{equation}
\label{eq:opt_cost}
J(s_i^t, \tilde{s}_j^{t-1}, \mathbf{b}_i^t, \widetilde{\mathbf{b}}_j^{t-1})
=
\mathbf{1}(s_i^t = \tilde{s}_j^{t-1}) \cdot \sum_{p=1}^{n} \delta_p(b_{i,p}^t, \tilde{b}_{j,p}^{t-1})
\end{equation}
Here, $\mathbf{1}(s_i^t = \tilde{s}_j^{t-1})$ is an indicator function.
$\delta_p(b_{i,p}^t, \tilde{b}_{j,p}^{t-1})$ is the 
consistency cost for the $p$-th attribute between segment $m_i^t$ in frame $t$ and track $T_j^{t-1}$. Importantly, one of these $\delta_p$ functions specifically accounts for the initial track IDs (\textit{c.f.}~\cref{eq:cost_s}), encouraging our optimization to maintain these associations when appropriate.

We use the Hungarian algorithm to solve for the new segment IDs and update the initial segment IDs only if the optimisation cost from \cref{eq:opt_cost} is below a cost threshold $\gamma$.
This ensures that our algorithm does not change associations when the cost is too high.
Notably, for new observations that don't match any existing track (i.e., their matching cost exceeds $\gamma$), we initialize new tracks.
Importantly, we do not terminate tracks that fail to match with a new observation in the current frame.
Instead, we maintain these tracks in our database, propagating their attributes from time $t-1$ to time $t$.
This approach allows our method to handle temporary occlusions or brief disappearances of objects, maintaining object identity over longer periods.

By iteratively applying this process across the entire video sequence, we refine the initial tracks, correcting errors while still leveraging the valuable information provided by the 2D VOS model.
Our method's ability to incorporate both the initial 2D tracking information and additional 3D cues, combined with its frame-by-frame processing and track maintenance strategy, enables it to effectively handle the challenges of egocentric videos, including frequent occlusions, objects moving in and out of view, and rapid camera motion.
Next, we describe how we define and compute the segment attributes $b_{i}^t$ as well as the associated cost functions $\delta_p$.

\subsection{Attributes for 3D-aware cost formulation}%
\label{sec:attributes}

Our method leverages 3D information to improve the \textit{initial} object tracks obtained from an off-the-shelf 2D-only VOS model.
In addition to 3D location information, we leverage appearance information (visual features), as well as categorical information (\ie, the initial category and instance labels from the 2D model) to refine the segment associations.
We denote the attributes for each segment as
$
\mathbf{b}_i^t
=
(l_{i}^t, v_{i}^t, c_{i}^t, s_{i}^t)
$,
where $l_{i}^t$ is the 3D location of the segment, $v_{i}^t$ is the visual feature, $c_{i}^t$ is the category label and $s_{i}^t$ is the instance label.

\paragraph{3D locations as segment attributes.}
We are given for each image $\mathbf{I}^t, t \in \{1, \dots, N\}$, a camera pose $\mathbf{C}^t$, camera intrinsics $K$ and a depth map $\mathbf{D}^t$.
In order to optimise the associations with 3D information, we lift the 2D centroid of each segment into 3D.
We define the 3D centroid of segment $m_i^t$ in frame $t$ as $l_{i}^t$, representing one out of several attributes of $\mathbf{b}_i^t$.
We calculate the location of this segment by projecting its 2D centroid into 3D with
\begin{equation}
    \label{eq:3d-feat}
    l_{i}^t = \mathbf{C}^t \left[ \begin{matrix} d_i^t K^{-1} \left[ \begin{matrix} x_i^t, & y_i^t, & 1 \end{matrix} \right]^{T} \\ 1 \end{matrix} \right],
\end{equation}
where $d_i^t$ is the depth value obtained from $\mathbf{D}^t$ that corresponds to the centroid of segment $m_i^t$ of frame $t$, and $x_i^t, y_i^t$ are the 2D coordinates of the centroid.

\paragraph{Visual features as segment attributes.}

While the 3D location of a segment plays a crucial role in overcoming the mentioned problems of associating segments throughout occlusions, viewpoint changes and similar issues, we also make use of 2D-level visual features $v_i^t$ as one of the attributes $\mathbf{b}_i^t$ that correspond to each segment.
Specifically, for an image $\mathbf{I}^t$ and each segment $m_i^t$ of the image, we use a pretrained vision encoder, \eg DINOv2~\cite{oquab2023dinov2}, to obtain the visual feature $v_i^t$ as:
\begin{equation}
    \label{eq:vis-feat}
    v_i^t = V(\text{crop}(\mathbf{I}^t \odot m_i^t)),
\end{equation}
where $V$ is the vision encoder and $\odot$ denotes Hadamard product.
The `crop' operation extracts the smallest patch with a 1:1 aspect ratio enclosing mask $m_i^t$.

\paragraph{Initial instance and category labels as segment attributes.}%
\label{sec:3d-aware}

Our proposed method refines the \textit{initial} tracks obtained from a purely 2D video object segmentation model.
Let $\bar{c}_i^t$ and $\bar{s}_i^t$ denote the initial category and instance labels for segment $m_i^t$ obtained from the 2D model.
We use $\bar{c}_i^t$ as an attribute to discourage the optimisation from matching instances which did not initially belong to the same category.
And similarly, we use $\bar{s}_i^t$ to encourage the optimization to preserve the initial tracks of instances across frames obtained from the 2D model.
We mathematically define the associated costs below.

\paragraph{Attributes for a track.}%
\label{sec:track-attr}

A track $\mathbf{T}^{t-1}$ that exists at time $t-1$ is a sequence of segments assigned to it so far.
We associate each track with an attribute vector $\widetilde{\mathbf{b}}_i^{t-1}=(\tilde{l}_{i}^t, \tilde{v}_{i}^t, \tilde{c}_{i}^t, \tilde{s}_{i}^t)$, where $\tilde{l}_{i}^t$, $\tilde{c}_{i}^t$ and $\tilde{s}_{i}^t$ are defined to be the corresponding attributes of the most recent segment assigned to this track. The visual feature attribute $\tilde{v}_{i}^t$ is defined to be the mean visual feature of the $100$ most recent segments assigned to the track.

\subsection{Cost functions}%
\label{sec:cost-matching}

The attributes used for refining the tracks are thus $\mathbf{b}_i^t = (l_{i}^t, v_{i}^t, \bar{c}_{i}^t, \bar{s}_{i}^t)$, consisting of the 3D location, the visual features, \textit{initial} category label and \textit{initial} instance label for the segment $m_i^t$ of frame $t$.
Now, we define the cost functions $\delta_p$ used in \cref{eq:opt_cost} for these individual attributes. 
We follow~\cite{Rajasegaran_2022_CVPR,plizzari2023} for the first two:
\begin{enumerate}
    \item We model the 3D location cost $\delta_l$ with the exponential distribution as follows:
    \begin{equation}
    \label{eq:cost_l}
    \delta_l(l_{i}^t, l_{j}^{t'}) = -\log\left(\frac{1}{\alpha_l} \exp\left(-\| l_i^t - l_j^{t'} \|_2 \right)\right)
    \end{equation}
    \item We model the cost for the visual features, $\delta_v$, using a Cauchy distribution:
    \begin{equation}
    \label{eq:cost_v}
    \delta_v(v_{i}^t, v_{j}^{t'}) = -\log\left(\frac{1}{1 + \alpha_v \| v_i^t - v_j^{t'} \|_2^2}\right)
    \end{equation}
    \item For the category and instance label, we use a $0-1$ cost function and refer to it with $\delta_c$ and $\delta_s$:
    
    \begin{minipage}{.5\linewidth}
    \begin{equation}
        \label{eq:cost_c}
        \delta_c(\bar{c}_{i}^t, \bar{c}_{j}^{t'})  = \begin{cases}
        0 & \text{if } \bar{c}_{i}^t = \bar{c}_{j}^{t'} \\
        \alpha_c & \text{if } \bar{c}_{i}^t \neq \bar{c}_{j}^{t'}
        \end{cases},
    \end{equation}
    \end{minipage}%
    \begin{minipage}{.5\linewidth}
    \begin{equation}
        \label{eq:cost_s}
        \delta_s(\bar{s}_{i}^t, \bar{s}_{j}^{t'})  = \begin{cases}
        0 & \text{if } \bar{s}_{i}^t = \bar{s}_{j}^{t'} \\
        \alpha_s & \text{if } \bar{s}_{i}^t \neq \bar{s}_{j}^{t'}
        \end{cases}
    \end{equation}
    \end{minipage}
\end{enumerate}
Here, $\alpha_l, \alpha_v$, $\alpha_c$ and $\alpha_s$ are used to modulate the importance of each cost function.
The cost parameters for the category and instance labels discourage the matching of segments that are inconsistent with the category and instance labels from the input segments.
As described in \Cref{sec:algorithm}, we consider the tracks formed in previous $t-1$ frames and match them to the new observations from the current frame $t$ using the Hungarian algorithm.

We refer the reader to \Cref{sec:impl} for the implementation details and to \Cref{sec:hyperparam} for hyperparameter settings.

\section{Experiments}%
\label{sec:experiments}

\subsection{Benchmark and baselines}%
\label{sec:baselines}

We evaluate our proposed method on 20 challenging
scenes from the EPIC Fields~\cite{EPICFields2023} dataset.
EPIC Fields comprises of complex real-world videos with a high diversity of activities and object interactions, making it an ideal testbed for our evaluation.
The selected videos include varied lighting conditions, occlusions, objects that disappear from sight, and have an average length of 10 minutes.
To further demonstrate our method's capability, we also evaluate it on the Ego4D \cite{grauman2022ego4d} dataset and report the results in \Cref{tab:ego4d}.

We compare against the following baselines:
\begin{enumerate}
    \item \textbf{DEVA}~\cite{cheng2023tracking} employs a decoupled video segmentation approach that combines task-specific image-level segmentation with a class-agnostic bi-directional temporal propagation model.
    This method is particularly effective in diverse and data-scarce environments, as it separates image and video segmentation tasks to improve overall tracking accuracy by reducing the impact of image segmentation errors.
    \item \textbf{MASA}~\cite{masa} is a more recent state-of-the-art method that focuses on robust instance association learning.
    MASA includes a universal adapter that allows it to integrate with various foundational segmentation or detection models, enhancing its ability to track any detected objects robustly.
    By utilizing features from these underlying 2D models, MASA can improve the instance and category assignments, providing robust zero-shot tracking capabilities in complex domains.
\end{enumerate}
Note that, both DEVA and MASA can be used with various 2D object detection models.
We tested both methods with three 2D models: OWLv2~\cite{owlv2}, Detic~\cite{zhou2022detecting} and GroundingDINO~\cite{liu2023grounding}, and found that DEVA works best with OWLv2 while MASA works best with Detic on the EPIC Fields dataset.
Hence, we incorporate \textbf{DEVA + OWLv2} and \textbf{MASA + Detic} as baselines in our experiments.

Since both baselines use an open-vocabulary 2D detection model, we use text prompts corresponding to the object categories from EPIC Fields~\cite{EPICFields2023} to obtain image-level object bounding boxes (with associated class labels).

In \Cref{sec:osnom}, we carry out an ablation disabling the category and instance terms in our cost function. This brings our approach somewhat closer to 
the method of OSNOM~\cite{plizzari2023}.


\subsection{Metrics}%
\label{sec:metrics}

We evaluate our method using the HOTA (Higher Order Tracking Accuracy) metric~\cite{luiten2021hota}.
HOTA assesses multi-object tracking (MOT) performance by combining detection accuracy (DetA), association accuracy (AssA), and localization IoU (Loc-IoU).
It is calculated as the geometric mean of DetA and AssA over various Loc-IoU thresholds $\alpha$:
\[
\text{HOTA}
= \frac{1}{|S|} \sum_{\alpha \in S} \text{HOTA}(\alpha)
= \frac{1}{|S|} \sum_{\alpha \in S} \sqrt{
    \text{DetA}(\alpha) \times \operatorname{AssA}(\alpha)
}
\]
where \(S\) is the set of IoU thresholds.
We use $S=\{0.05,0.1,\dots,0.9,0.95\}$ following standard protocol~\cite{luiten2021hota}.
DetA measures the overlap between the set of \textit{all} predicted segments and \textit{all} ground-truth (GT) segments.
It is defined as:
\[
\text{DetA}(\alpha)
=
\frac
{|\mathrm{TP}_{\alpha}|}
{|\mathrm{TP}_{\alpha}|+|\text{FP}_{\alpha}|+|\text{FN}_{\alpha}|}
\]
True Positives (TP${_\alpha}$) are identified by matching predicted segments to GT segments with an IoU $\geq \alpha$ using Hungarian matching.
Unmatched predictions are False Positives (FP$_{\alpha}$), and unmatched GT segments are False Negatives (FN$_{\alpha}$).

AssA measures the tracker's ability to maintain consistent object identities over time:
\[
\operatorname{AssA}(\alpha)
= 
\frac{1}{|\mathrm{TP}_{\alpha}|} 
\sum_{c \in \mathrm{TP}_{\alpha}}
\frac
{\operatorname{TPA}(c)|}
{|\operatorname{TPA}(c)| + |\operatorname{FPA}(c)| + |\operatorname{FNA}(c)|}
\]
where we iterate over all TP pairs, measuring the alignment between the predicted and ground-truth segment's \textit{whole} track.
True Positive Associations (TPA) represents the number of TP matches between the two chosen tracks for a pair.

Additionally, we use the IDF1 (Identity F1) score to measure how well the tracker maintains consistent object identities throughout the sequence:
\[
\mathrm{IDF1}
=
\frac
{2 \, |\mathrm{IDTP}|}
{2 \, |\mathrm{IDTP}| + |\text{IDFP}| + |\text{IDFN}|}
\]
where IDTP (Identity True Positives) represents matches on overlapping parts of tracks that are matched, while IDFP (Identity False Positives) and IDFN (Identity False Negatives) represent the remaining GT and predicted segments.

\subsection{Results}%
\label{sec:results}

\begin{figure}[!ht]
    \centering
    \includegraphics[width=1.0\textwidth]{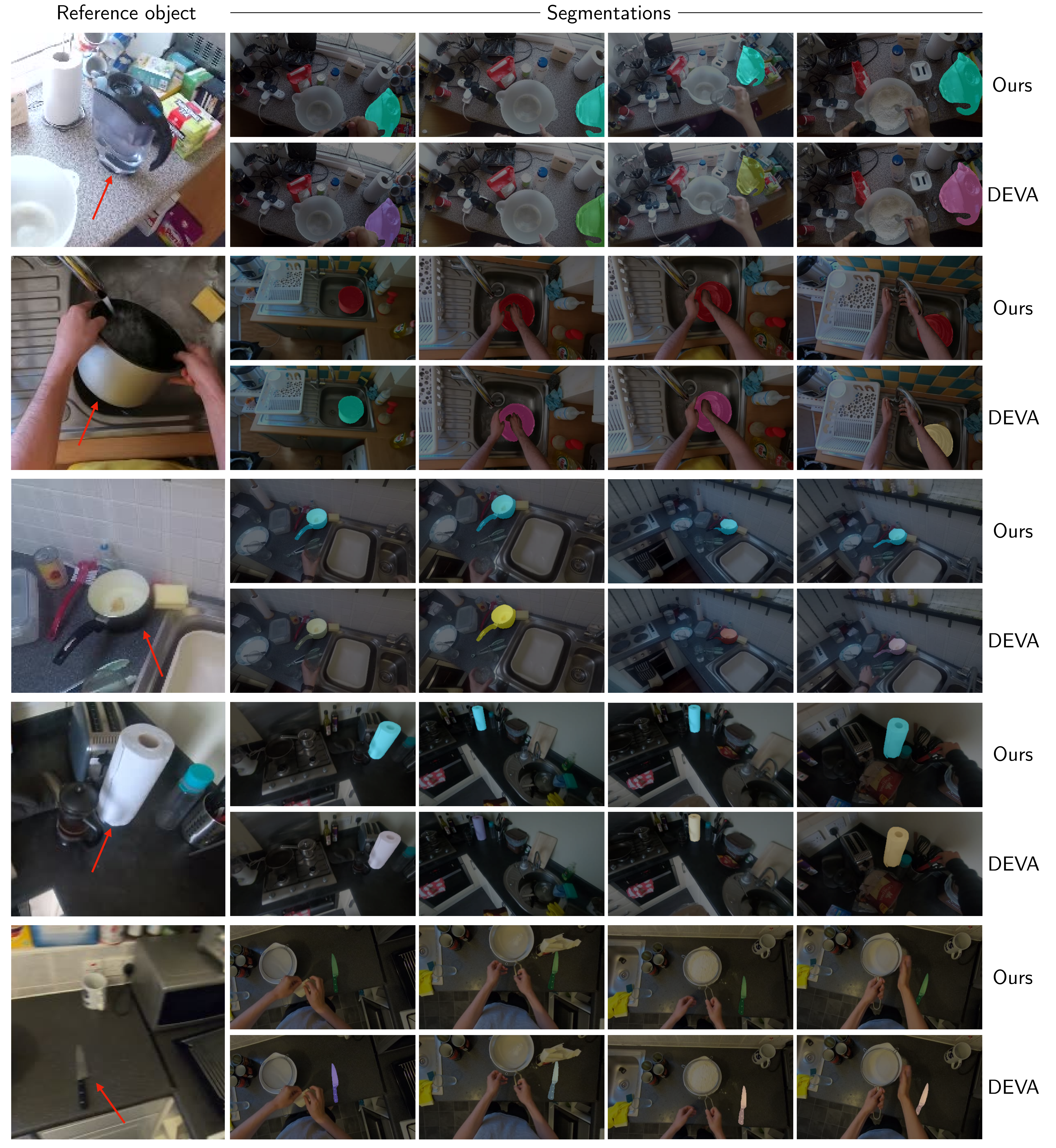}
    \caption{Qualitative comparison between our method and DEVA~\cite{cheng2023tracking}.
    We show instance segmentations for selected reference objects.
    Our method maintains consistent tracks despite viewpoint changes and objects going out of view, while DEVA's tracks break.
    Our approach successfully segments the pot even when in motion.}%
    \label{fig:qualitative}
\end{figure}

We evaluate our method against DEVA~\cite{cheng2023tracking} and MASA~\cite{masa} using the HOTA, DetA, AssA, and IDF1 metrics.
\Cref{tab:quant-results} presents the overall results as well as scene-specific performance. \Cref{fig:qualitative} provides a qualitative comparison of results.

Our approach consistently outperforms both baselines across various metrics.
Compared to DEVA, our method achieves an overall HOTA score of 27.72, a notable improvement over DEVA's 25.14.
This enhancement is even more pronounced in the AssA metric, which measures the tracker's ability to maintain consistent object identities over time.
Our method attains an AssA score of 43.90, substantially higher than DEVA's 36.72.

This further underscores our method's superior performance in maintaining consistent object identities throughout the video sequences.
Our method also shows significant improvements in IDF1 scores, achieving 26.63 compared to DEVA's 22.17.
Similar improvements are observed when comparing to MASA, which demonstrates our approach's adaptability to different base models.

Notably, DetA scores remain relatively consistent across all methods (\eg 18.40 for MASA vs.~18.38 for our method when using MASA as the base model).
This is because our method improves the instance and category assignments for the segments using 3D information but does not alter the segments themselves.
Since the DetA metric only evaluates the segments regardless of IDs, it results in similar scores for both the base 2D method and our method.

\paragraph{Scene-specific analysis.}

Our method shows remarkable improvements in complex scenes, such as P01\_01, where we achieve a HOTA score of 41.91 compared to DEVA's 33.60, a 24\% improvement.
This scene likely contains frequent object occlusions or out-of-view instances where our 3D-aware approach excels.
Significant improvements are also observed in scenes like P07\_101 and P22\_117, with improvements of 25\% and 22\% respectively in HOTA scores.

The AssA metric shows the most significant improvements.
For example, in P02\_121, our method achieves an AssA of 20.96 compared to DEVA's 10.32, a 103\% improvement.
However, the degree of improvement varies across scenes.
In some, like P04\_11, the improvement is marginal, suggesting that not all scenes benefit equally from 3D awareness.

\begin{table}[t]
\centering
\caption{Results on the EPIC Fields~\cite{EPICFields2023} dataset.}%
\label{tab:quant-results}
\newcolumntype{d}{D{.}{.}{2}}
\resizebox{\textwidth}{!}{
\begin{tabular}{l cccc  cccc cccc  cccc}
\toprule
\multirow{2}{*}{Video} & \multicolumn{4}{c}{DEVA~\cite{cheng2023tracking}} & \multicolumn{4}{c}{Ours (w/ DEVA)} & \multicolumn{4}{c}{MASA~\cite{masa}} & \multicolumn{4}{c}{Ours (w/ MASA)}\\
\cmidrule(r){2-5}
\cmidrule(lr){6-9}
\cmidrule(lr){10-13}
\cmidrule(l){14-17}
& HOTA  & DetA  & AssA  & IDF1  & HOTA  & DetA  & AssA  & IDF1  & HOTA  & DetA  & AssA  & IDF1  & HOTA  & DetA  & AssA  & IDF1 \\
\midrule
P01\_01  & 33.60 & 25.25 & 45.68 & 28.61 & 41.91 & 24.94 & 71.85 & 38.76 & 9.11  & 4.64  & 17.99 & 8.15  & 8.36  & 4.64  & 15.12 & 7.50  \\
P01\_104 & 25.79 & 22.98 & 29.09 & 21.93 & 30.92 & 22.95 & 41.88 & 31.40 & 11.66 & 8.81  & 15.59 & 9.61  & 12.77 & 8.81  & 18.64 & 10.54 \\
P02\_09  & 30.07 & 21.85 & 42.29 & 23.46 & 33.76 & 21.77 & 53.11 & 27.73 & 20.51 & 15.46 & 27.46 & 17.67 & 19.04 & 15.47 & 23.68 & 16.45 \\
P02\_121 & 8.75  & 7.47  & 10.32 & 6.07  & 11.79 & 6.64  & 20.96 & 12.09 & 6.71  & 5.68  & 8.03  & 4.06  & 9.29  & 5.69  & 15.34 & 6.90  \\
P02\_132 & 26.71 & 25.05 & 28.80 & 29.04 & 29.96 & 24.74 & 36.56 & 35.18 & 15.44 & 11.40 & 21.28 & 13.31 & 15.39 & 11.35 & 20.98 & 13.65 \\
P03\_101 & 27.56 & 21.07 & 36.13 & 24.17 & 29.63 & 19.72 & 44.61 & 26.67 & 7.71  & 6.22  & 9.65  & 4.55  & 9.53  & 6.22  & 14.76 & 6.97  \\
P04\_03  & 15.60 & 11.72 & 23.41 & 11.24 & 16.85 & 11.64 & 26.63 & 12.23 & 10.21 & 5.12  & 22.80 & 6.22  & 10.17 & 5.12  & 21.92 & 6.67  \\
P04\_11  & 43.03 & 35.83 & 52.05 & 48.88 & 43.13 & 35.87 & 52.21 & 49.74 & 10.82 & 7.26  & 16.30 & 11.26 & 10.54 & 7.27  & 15.37 & 9.76  \\
P04\_25  & 18.71 & 6.02  & 58.25 & 10.39 & 18.71 & 6.18  & 56.79 & 10.64 & 12.64 & 5.96  & 27.30 & 6.69  & 13.46 & 5.96  & 30.45 & 8.54  \\
P06\_01  & 26.22 & 23.80 & 29.60 & 28.12 & 29.73 & 25.65 & 34.95 & 35.05 & 18.95 & 21.33 & 19.02 & 17.96 & 26.01 & 21.33 & 32.87 & 30.84 \\
P06\_102 & 27.71 & 17.37 & 44.81 & 23.75 & 30.42 & 18.09 & 51.88 & 28.71 & 10.42 & 6.17  & 18.87 & 4.87  & 8.71  & 6.18  & 13.91 & 4.17  \\
P06\_12  & 42.47 & 27.00 & 68.89 & 41.40 & 44.13 & 26.95 & 73.86 & 48.41 & 41.94 & 28.57 & 62.01 & 47.70 & 44.35 & 28.54 & 69.14 & 52.38 \\
P07\_101 & 18.45 & 15.66 & 23.28 & 14.28 & 23.12 & 15.95 & 34.81 & 21.44 & 12.25 & 7.83  & 19.98 & 8.58  & 12.98 & 7.82  & 22.38 & 9.44  \\
P11\_103 & 27.73 & 15.55 & 49.78 & 24.77 & 24.68 & 15.16 & 40.42 & 21.98 & 11.69 & 8.11  & 17.37 & 7.75  & 13.25 & 8.02  & 21.98 & 9.57  \\
P12\_02  & 23.51 & 15.26 & 37.40 & 16.16 & 26.21 & 15.45 & 45.35 & 20.63 & 11.46 & 7.33  & 17.96 & 7.46  & 12.77 & 7.34  & 22.34 & 9.59  \\
P22\_117 & 18.15 & 12.62 & 27.06 & 13.50 & 22.06 & 12.34 & 39.90 & 18.92 & 7.46  & 3.32  & 17.88 & 4.63  & 6.29  & 3.33  & 12.37 & 3.97  \\
P24\_05  & 19.07 & 12.27 & 30.48 & 16.10 & 21.02 & 12.27 & 36.50 & 19.85 & 11.39 & 9.12  & 14.26 & 7.27  & 13.15 & 9.09  & 19.06 & 10.00 \\
P28\_109 & 24.77 & 17.39 & 35.36 & 21.68 & 25.99 & 18.08 & 37.37 & 26.32 & 12.82 & 11.49 & 14.38 & 9.97  & 13.29 & 11.49 & 15.41 & 10.97 \\
P28\_14  & 27.11 & 18.85 & 39.90 & 25.30 & 28.04 & 18.18 & 44.28 & 27.54 & 13.22 & 9.21  & 20.28 & 10.17 & 13.35 & 9.17  & 20.16 & 10.61 \\
P37\_101 & 17.85 & 14.97 & 21.79 & 14.56 & 22.23 & 14.93 & 33.96 & 19.24 & 11.09 & 9.26  & 13.86 & 8.54  & 11.20 & 9.14  & 14.07 & 8.76  \\
\midrule
Overall & 25.14 & 18.40 & 36.72 & 22.17 & \textbf{27.72} & 18.38 & \textbf{43.90} & \textbf{26.63} & 13.73 & 10.06 & 20.32 & 10.82 & \textbf{14.67} & 10.04 & \textbf{22.43} & \textbf{12.36} \\
\bottomrule
\end{tabular}}
\end{table}

\begin{table}[t]
\centering
\caption{Number of ID switches averaged over all videos, shown for challenging and frequently appearing objects. Last column: number of videos featuring each object.}%
\label{tab:id-switch}
\setlength{\tabcolsep}{6pt}
\begin{tabular}{lccc}
\toprule
Object Class & DEVA~\cite{cheng2023tracking} & Ours (w/ DEVA) & \# videos\\
\midrule
\textit{tap} & 14.53 & \textbf{2.88}  & 17 \\
\textit{knife} & 27.21 & \textbf{5.29} & 14 \\
\textit{chopping board} & 20.25 & \textbf{5.42} & 12  \\
\textit{spoon} & 21.00 & \textbf{5.00} & 10 \\
\textit{bowl} & 23.11 & \textbf{5.67} & 9  \\
\textit{pan} & 19.44 & \textbf{4.11} & 9  \\
\textit{sponge} & 22.38 & \textbf{5.38} & 8  \\
\bottomrule
\end{tabular}
\end{table}

\paragraph{Analysis of ID switches by object class.}

To further understand our method's performance in maintaining consistent object identities, we analyze the number of ID switches occurring throughout the videos for different object categories.
\Cref{tab:id-switch} shows the average number of ID switches over all videos for a subset of challenging and commonly occurring object classes in the EPIC Fields dataset, comparing our method to the DEVA baseline.
Our approach consistently and significantly reduces ID switches across all shown object classes, with improvements ranging from 73\% to 80\% reduction.
For instance, small objects, prone to occlusions, such as knives, see a reduction from 27.21 to 5.29 switches, taps from 14.53 to 2.88, and pans from 19.44 to 4.11.
This substantial improvement across various object types, regardless of their size or frequency of appearance, demonstrates the robustness of our 3D-aware approach.
It highlights our method's effectiveness in maintaining consistent object identities through complex interactions and occlusions typical in egocentric videos, particularly for frequently manipulated kitchen objects and objects that may remain stationary across time, while not necessarily staying in view.

\subsection{Ablations}%
\label{sec:ablations}

\paragraph{Comparison with other \textit{plug-and-play} tracking methods}
The above results demonstrated our method's generalization capability by combining with with two state-of-the-art methods, DEVA~\cite{cheng2023tracking} and MASA~\cite{masa}.
To further highlight our method's versatility, we compare it with other existing \textit{plug-and-play} tracking algorithms, namely BoTSORT~\cite{botsort}, ByteTrack~\cite{bytetrack}, OCSORT~\cite{ocsort} and DeepOCSORT~\cite{deepocsort}.
We use the same ReID model with all 4 tracking methods and use OWLv2~\cite{owlv2} as the 2D segmentation model for fair comparisons.
\Cref{tab:additional-comp} shows that all four methods perform less favourably than DEVA~\cite{cheng2023tracking} even while using the same base 2D model, and thus are outperformed by our method which further refines the tracks from DEVA.

\begin{table}[t]
\centering
\caption{Comparison with other plug-and-play tracking methods.}%
\label{tab:additional-comp}
\begin{tabular}{lcccc}
\toprule
Method & HOTA & DetA & AssA & IDF1 \\
\midrule
BoTSORT~\cite{botsort} & 12.83 & 7.81 & 24.23 & 10.30 \\
ByteTrack~\cite{bytetrack} & 20.08 & 16.31 & 27.56 & 16.94 \\
OCSORT~\cite{ocsort} & 21.90 & 17.94 & 29.28 & 18.95 \\
DeepOCSORT~\cite{deepocsort} & 22.63 & 17.98 & 31.31 & 19.88 \\
\midrule
DEVA~\cite{cheng2023tracking}  & 25.14 & \textbf{18.40} & 36.72 & 22.17 \\
\textbf{Ours} (w/ DEVA) & \textbf{27.72} & 18.38 & \textbf{43.90} & \textbf{26.63} \\
\bottomrule
\end{tabular}
\end{table}

\paragraph{Influence of different components on tracking.}

Our tracking formulation consists of four components (\cref{eq:cost_s,eq:cost_c,eq:cost_l,eq:cost_v}): instance cost, category cost, 3D location cost, and visual feature cost.
We evaluate the influence of each component by turning off the corresponding cost one at a time in the cost-matching formulation.
\Cref{tab:ablation-attributes} shows that all components contribute positively to the tracking performance, but to varying degrees.
Removing the visual features has the least impact, reducing the HOTA score from 27.72 to 27.17.
The 3D location information proves more important, with its removal causing the HOTA score to drop to 26.32.
Removing the category term has the most significant impact on the tracking performance, followed by the instance cost.
Note that, if the instance cost is removed, the cost optimization completely ignores the \textit{initial} tracks provided by the 2D base tracker (\eg DEVA or MASA), effectively finding instance tracks from scratch.
Notably, even without this initial guidance, our method outperforms the 2D tracking method (DEVA~\cite{cheng2023tracking}) in terms of HOTA ($+0.82$), AssA ($+2.79$) and IDF1 ($+2.33$).

\begin{table}[t]
\centering
\caption{Influence of different components in the tracking formulation.}%
\label{tab:ablation-attributes}
\setlength{\tabcolsep}{6pt}
\begin{tabular}{cccc cccc}
    \toprule
    {Instance} & {Category} & {3D Location} & {Visual} & HOTA    & DetA  & AssA    & IDF1    \\
    \cmidrule[0.5pt](r){1-4}
    \cmidrule[0.5pt](l){5-8}
    \cmark     & \cmark     & \cmark        & \cmark   & {27.72} & 18.38 & {43.90} & {26.63} \\
    \cmark     & \cmark     & \cmark        & \xmark   & 27.17   & 18.38 & 42.45   & 26.12   \\
    \cmark     & \cmark     & \xmark        & \cmark   & 26.32   & 18.37 & 41.23   & 26.04   \\
    \cmark     & \xmark     & \cmark        & \cmark   & 25.49   & 18.11 & 38.74   & 24.18   \\
    \xmark     & \cmark     & \cmark        & \cmark   & 25.96   & 18.38 & 39.51   & 24.50   \\
    \xmark     & \cmark     & \xmark        & \xmark   & 21.11 & 18.41 & 26.42 & 16.80 \\
    \cmark     & \cmark     & \xmark        & \xmark   & 25.14 & 18.40 & 36.72 & 22.19 \\
    \midrule
    \multicolumn{4}{c}{DEVA~\cite{cheng2023tracking}} & 25.14 & 18.40 & 36.72 & 22.17 \\
    \bottomrule
\end{tabular}
\end{table}

\paragraph{Metrics across IoU thresholds.}

\begin{figure}[t]
    \centering
    \includegraphics[width=0.95\textwidth]{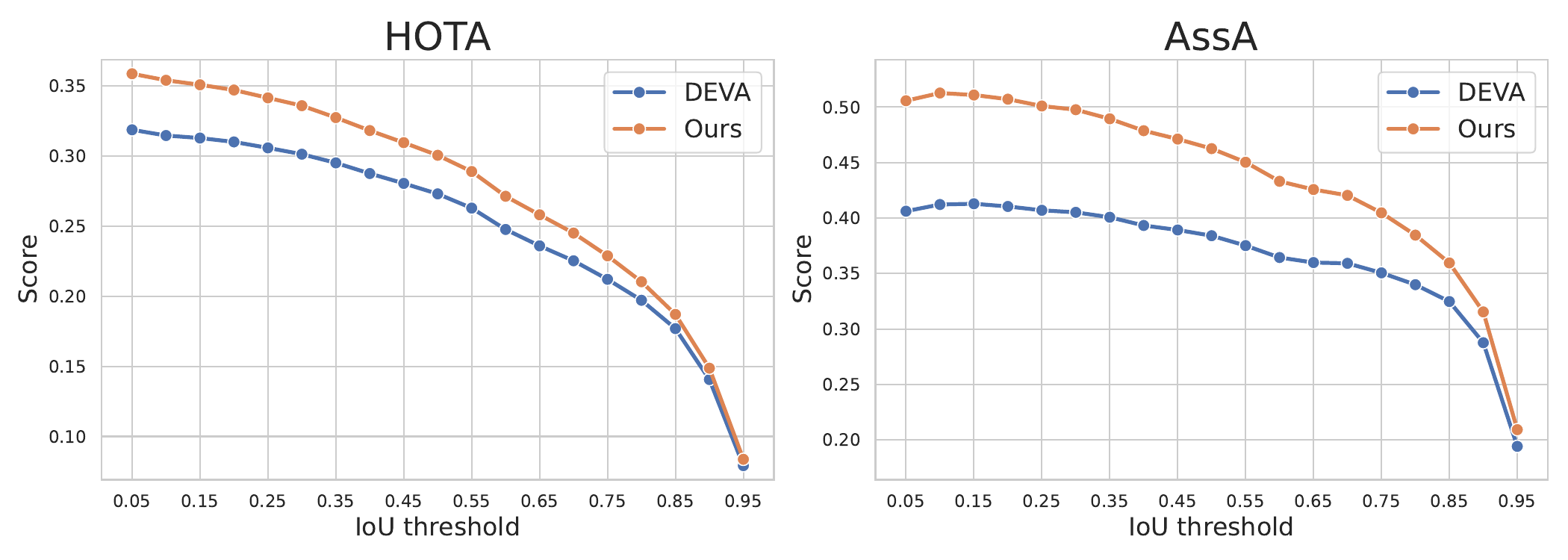}
    \vspace{-1em}
    \caption{HOTA and Association accuracy (AssA) metrics across different IoU thresholds.}%
    \label{fig:thresh-metrics}
\end{figure}

As described in \Cref{sec:metrics}, HOTA, DetA, and AssA can be calculated at different IoU thresholds.
\Cref{fig:thresh-metrics} illustrates how these metrics change as the IoU threshold increases.
As expected, all metrics decrease with higher thresholds, as stricter overlap requirements lead to fewer True Positive matches between predicted and ground-truth segments.
Notably, our method consistently outperforms DEVA across all thresholds for both HOTA and AssA metrics, while he AssA curve shows a more pronounced improvement.
This suggests that our 3D-aware approach is particularly effective at maintaining consistent object identities throughout the video sequence, even under strict evaluation criteria.

\section{Conclusion}%
\label{sec:conclusion}

In this paper, we presented a novel 3D-aware approach to instance segmentation and tracking in egocentric videos, addressing the unique challenges of first-person perspectives. By integrating 3D information, our method significantly improves tracking accuracy and segmentation consistency compared to state-of-the-art 2D approaches, especially over long periods.
Our ablation studies highlight the importance of 3D information and the category as well as instance cost terms in matching, while also showing robustness to hyperparameter changes.
Beyond improved tracking, our approach enables valuable downstream applications such as high-quality 3D object reconstructions and amodal segmentation. This work demonstrates the power of incorporating 3D awareness into egocentric video analysis, opening up new possibilities for robust object tracking in challenging first-person scenarios.

\begin{credits}
\subsubsection{\ackname}
We're funded by EPSRC AIMS CDT EP/S024050/1 and AWS (Y.~Bhalgat), NAVER LABS Europe (V.~Tschernezki), ERC-CoG UNION 101001212 (A.~Vedaldi and I.~Laina), EPSRC VisualAI EP/T028572/1 (I.~Laina, A.~Vedaldi and A.~Zisserman), and Royal Academy of Engineering RF\textbackslash 201819\textbackslash 18\textbackslash 163 (J.~Henriques).
We thank Chiara Plizzari for sharing evaluation details of the OSNOM baseline, and Ahmad Dar Khalil for helping with the annotations for our evaluation.
\end{credits}

\bibliographystyle{splncs04}
\bibliography{main}

\begin{thebibliography}{10}
\providecommand{\url}[1]{\texttt{#1}}
\providecommand{\urlprefix}{URL }
\providecommand{\doi}[1]{https://doi.org/#1}

\bibitem{botsort}
Aharon, N., Orfaig, R., Bobrovsky, B.Z.: Bot-sort: Robust associations multi-pedestrian tracking. arXiv preprint arXiv:2206.14651  (2022)

\bibitem{athar2023burst}
Athar, A., Luiten, J., Voigtlaender, P., Khurana, T., Dave, A., Leibe, B., Ramanan, D.: Burst: A benchmark for unifying object recognition, segmentation and tracking in video. In: Proceedings of the IEEE/CVF winter conference on applications of computer vision. pp. 1674--1683 (2023)

\bibitem{bhalgat2024contrastive}
Bhalgat, Y., Laina, I., Henriques, J.F., Vedaldi, A., Zisserman, A.: Contrastive lift: 3d object instance segmentation by slow-fast contrastive fusion. Advances in Neural Information Processing Systems  \textbf{36} (2024)

\bibitem{bhalgat2024n2f2}
Bhalgat, Y., Laina, I., Henriques, J.F., Zisserman, A., Vedaldi, A.: N2f2: Hierarchical scene understanding with nested neural feature fields. arXiv preprint arXiv:2403.10997  (2024)

\bibitem{davis2019}
Caelles, S., Pont-Tuset, J., Perazzi, F., Montes, A., Maninis, K.K., Van~Gool, L.: The 2019 davis challenge on vos: Unsupervised multi-object segmentation. arXiv preprint arXiv:1905.00737  (2019)

\bibitem{ocsort}
Cao, J., Pang, J., Weng, X., Khirodkar, R., Kitani, K.: Observation-centric sort: Rethinking sort for robust multi-object tracking. In: Proceedings of the IEEE/CVF Conference on Computer Vision and Pattern Recognition. pp. 9686--9696 (2023)

\bibitem{chen2024panoptic}
Chen, H., Blomqvist, K., Milano, F., Siegwart, R.: Panoptic vision-language feature fields. IEEE Robotics and Automation Letters  (2024)

\bibitem{cheng2023tracking}
Cheng, H.K., Oh, S.W., Price, B., Schwing, A., Lee, J.Y.: Tracking anything with decoupled video segmentation. In: Proceedings of the IEEE/CVF International Conference on Computer Vision. pp. 1316--1326 (2023)

\bibitem{cheng2022xmem}
Cheng, H.K., Schwing, A.G.: Xmem: Long-term video object segmentation with an atkinson-shiffrin memory model. In: European Conference on Computer Vision. pp. 640--658. Springer (2022)

\bibitem{choudhuri2021assignment}
Choudhuri, A., Chowdhary, G., Schwing, A.G.: Assignment-space-based multi-object tracking and segmentation. In: Proceedings of the IEEE/CVF International Conference on Computer Vision. pp. 13598--13607 (2021)

\bibitem{choudhuri2023context}
Choudhuri, A., Chowdhary, G., Schwing, A.G.: Context-aware relative object queries to unify video instance and panoptic segmentation. In: Proceedings of the IEEE/CVF Conference on Computer Vision and Pattern Recognition. pp. 6377--6386 (2023)

\bibitem{Damen2021PAMI}
Damen, D., Doughty, H., Farinella, G.M., Fidler, S., Furnari, A., Kazakos, E., Moltisanti, D., Munro, J., Perrett, T., Price, W., Wray, M.: The epic-kitchens dataset: Collection, challenges and baselines. IEEE Transactions on Pattern Analysis and Machine Intelligence (TPAMI)  \textbf{43}(11),  4125--4141 (2021). \doi{10.1109/TPAMI.2020.2991965}

\bibitem{doersch2022tap}
Doersch, C., Gupta, A., Markeeva, L., Recasens, A., Smaira, L., Aytar, Y., Carreira, J., Zisserman, A., Yang, Y.: Tap-vid: A benchmark for tracking any point in a video. Advances in Neural Information Processing Systems  \textbf{35},  13610--13626 (2022)

\bibitem{doersch2023tapir}
Doersch, C., Yang, Y., Vecerik, M., Gokay, D., Gupta, A., Aytar, Y., Carreira, J., Zisserman, A.: Tapir: Tracking any point with per-frame initialization and temporal refinement. In: Proceedings of the IEEE/CVF International Conference on Computer Vision. pp. 10061--10072 (2023)

\bibitem{grauman2022ego4d}
Grauman, K., Westbury, A., Byrne, E., Chavis, Z., Furnari, A., Girdhar, R., Hamburger, J., Jiang, H., Liu, M., Liu, X., et~al.: Ego4d: Around the world in 3,000 hours of egocentric video. In: Proceedings of the IEEE/CVF Conference on Computer Vision and Pattern Recognition. pp. 18995--19012 (2022)

\bibitem{gu2024egolifter}
Gu, Q., Lv, Z., Frost, D., Green, S., Straub, J., Sweeney, C.: Egolifter: Open-world 3d segmentation for egocentric perception. arXiv preprint arXiv:2403.18118  (2024)

\bibitem{harley2022particle}
Harley, A.W., Fang, Z., Fragkiadaki, K.: Particle video revisited: Tracking through occlusions using point trajectories. In: European Conference on Computer Vision. pp. 59--75. Springer (2022)

\bibitem{Huang2DGS2024}
Huang, B., Yu, Z., Chen, A., Geiger, A., Gao, S.: 2d gaussian splatting for geometrically accurate radiance fields. In: SIGGRAPH 2024 Conference Papers. Association for Computing Machinery (2024). \doi{10.1145/3641519.3657428}

\bibitem{karaev2023cotracker}
Karaev, N., Rocco, I., Graham, B., Neverova, N., Vedaldi, A., Rupprecht, C.: Cotracker: It is better to track together. arXiv preprint arXiv:2307.07635  (2023)

\bibitem{kerbl3Dgaussians}
Kerbl, B., Kopanas, G., Leimk{\"u}hler, T., Drettakis, G.: 3d gaussian splatting for real-time radiance field rendering. ACM Transactions on Graphics  \textbf{42}(4) (July 2023), \url{https://repo-sam.inria.fr/fungraph/3d-gaussian-splatting/}

\bibitem{kim2024garfield}
Kim, C.M., Wu, M., Kerr, J., Goldberg, K., Tancik, M., Kanazawa, A.: Garfield: Group anything with radiance fields. In: CVPR. pp. 21530--21539 (2024)

\bibitem{kirillov2023segment}
Kirillov, A., Mintun, E., Ravi, N., Mao, H., Rolland, C., Gustafson, L., Xiao, T., Whitehead, S., Berg, A.C., Lo, W.Y., et~al.: Segment anything. In: Proceedings of the IEEE/CVF International Conference on Computer Vision. pp. 4015--4026 (2023)

\bibitem{masa}
Li, S., Ke, L., Danelljan, M., Piccinelli, L., Segu, M., Van~Gool, L., Yu, F.: Matching anything by segmenting anything. CVPR  (2024)

\bibitem{liu2023grounding}
Liu, S., Zeng, Z., Ren, T., Li, F., Zhang, H., Yang, J., Li, C., Yang, J., Su, H., Zhu, J., et~al.: Grounding dino: Marrying dino with grounded pre-training for open-set object detection. arXiv preprint arXiv:2303.05499  (2023)

\bibitem{luiten2021hota}
Luiten, J., Osep, A., Dendorfer, P., Torr, P., Geiger, A., Leal-Taix{\'e}, L., Leibe, B.: Hota: A higher order metric for evaluating multi-object tracking. International journal of computer vision  \textbf{129},  548--578 (2021)

\bibitem{deepocsort}
Maggiolino, G., Ahmad, A., Cao, J., Kitani, K.: Deep oc-sort: Multi-pedestrian tracking by adaptive re-identification. In: 2023 IEEE International Conference on Image Processing (ICIP). pp. 3025--3029. IEEE (2023)

\bibitem{owlv2}
Matthias~Minderer, Alexey~Gritsenko, N.H.: Scaling open-vocabulary object detection. NeurIPS  (2023)

\bibitem{meinhardt2022trackformer}
Meinhardt, T., Kirillov, A., Leal-Taixe, L., Feichtenhofer, C.: Trackformer: Multi-object tracking with transformers. In: Proceedings of the IEEE/CVF conference on computer vision and pattern recognition. pp. 8844--8854 (2022)

\bibitem{oh2019video}
Oh, S.W., Lee, J.Y., Xu, N., Kim, S.J.: Video object segmentation using space-time memory networks. In: Proceedings of the IEEE/CVF international conference on computer vision. pp. 9226--9235 (2019)

\bibitem{oquab2023dinov2}
Oquab, M., Darcet, T., Moutakanni, T., Vo, H.V., Szafraniec, M., Khalidov, V., Fernandez, P., Haziza, D., Massa, F., El-Nouby, A., Howes, R., Huang, P.Y., Xu, H., Sharma, V., Li, S.W., Galuba, W., Rabbat, M., Assran, M., Ballas, N., Synnaeve, G., Misra, I., Jegou, H., Mairal, J., Labatut, P., Joulin, A., Bojanowski, P.: Dinov2: Learning robust visual features without supervision (2023)

\bibitem{davis2016}
Perazzi, F., Pont-Tuset, J., McWilliams, B., Van~Gool, L., Gross, M., Sorkine-Hornung, A.: A benchmark dataset and evaluation methodology for video object segmentation. In: Proceedings of the IEEE conference on computer vision and pattern recognition. pp. 724--732 (2016)

\bibitem{plizzari2023}
Plizzari, C., Goel, S., Perrett, T., Chalk, J., Kanazawa, A., Damen, D.: Spatial cognition from egocentric video: Out of sight, not out of mind. In: ArXiv (2024)

\bibitem{qiao2021vip}
Qiao, S., Zhu, Y., Adam, H., Yuille, A., Chen, L.C.: Vip-deeplab: Learning visual perception with depth-aware video panoptic segmentation. In: Proceedings of the IEEE/CVF Conference on Computer Vision and Pattern Recognition. pp. 3997--4008 (2021)

\bibitem{Rajasegaran_2022_CVPR}
Rajasegaran, J., Pavlakos, G., Kanazawa, A., Malik, J.: Tracking people by predicting 3d appearance, location and pose. In: Proceedings of the IEEE/CVF Conference on Computer Vision and Pattern Recognition (CVPR). pp. 2740--2749 (June 2022)

\bibitem{rajivc2023segment}
Raji{\v{c}}, F., Ke, L., Tai, Y.W., Tang, C.K., Danelljan, M., Yu, F.: Segment anything meets point tracking. arXiv preprint arXiv:2307.01197  (2023)

\bibitem{santrock2002topical}
Santrock, J.W.: A topical approach to life-span development. McGraw Hill (2002)

\bibitem{colmap}
Schonberger, J.L., Frahm, J.M.: Structure-from-motion revisited. In: Proceedings of the IEEE conference on computer vision and pattern recognition. pp. 4104--4113 (2016)

\bibitem{siddiqui2023panoptic}
Siddiqui, Y., Porzi, L., Bul{\'o}, S.R., M{\"u}ller, N., Nie{\ss}ner, M., Dai, A., Kontschieder, P.: Panoptic lifting for 3d scene understanding with neural fields. In: Proceedings of the IEEE/CVF Conference on Computer Vision and Pattern Recognition. pp. 9043--9052 (2023)

\bibitem{tang2024egotracks}
Tang, H., Liang, K.J., Grauman, K., Feiszli, M., Wang, W.: Egotracks: A long-term egocentric visual object tracking dataset. Advances in Neural Information Processing Systems  \textbf{36} (2024)

\bibitem{EPICFields2023}
Tschernezki, V., Darkhalil, A., Zhu, Z., Fouhey, D., Larina, I., Larlus, D., Damen, D., Vedaldi, A.: {EPIC Fields}: {M}arrying {3D} {G}eometry and {V}ideo {U}nderstanding. In: Proceedings of the Neural Information Processing Systems (NeurIPS) (2023)

\bibitem{voigtlaender2019feelvos}
Voigtlaender, P., Chai, Y., Schroff, F., Adam, H., Leibe, B., Chen, L.C.: Feelvos: Fast end-to-end embedding learning for video object segmentation. In: Proceedings of the IEEE/CVF conference on computer vision and pattern recognition. pp. 9481--9490 (2019)

\bibitem{wang2021end}
Wang, Y., Xu, Z., Wang, X., Shen, C., Cheng, B., Shen, H., Xia, H.: End-to-end video instance segmentation with transformers. In: Proceedings of the IEEE/CVF conference on computer vision and pattern recognition. pp. 8741--8750 (2021)

\bibitem{wu2022seqformer}
Wu, J., Jiang, Y., Bai, S., Zhang, W., Bai, X.: Seqformer: Sequential transformer for video instance segmentation. In: European Conference on Computer Vision. pp. 553--569. Springer (2022)

\bibitem{depthanything}
Yang, L., Kang, B., Huang, Z., Xu, X., Feng, J., Zhao, H.: Depth anything: Unleashing the power of large-scale unlabeled data. In: CVPR (2024)

\bibitem{yang2019video}
Yang, L., Fan, Y., Xu, N.: Video instance segmentation. In: Proceedings of the IEEE/CVF international conference on computer vision. pp. 5188--5197 (2019)

\bibitem{yang2021associating}
Yang, Z., Wei, Y., Yang, Y.: Associating objects with transformers for video object segmentation. Advances in Neural Information Processing Systems  \textbf{34},  2491--2502 (2021)

\bibitem{gaussian_grouping}
Ye, M., Danelljan, M., Yu, F., Ke, L.: Gaussian grouping: Segment and edit anything in 3d scenes. In: ECCV (2024)

\bibitem{ying2024omniseg3d}
Ying, H., Yin, Y., Zhang, J., Wang, F., Yu, T., Huang, R., Fang, L.: Omniseg3d: Omniversal 3d segmentation via hierarchical contrastive learning. In: Proceedings of the IEEE/CVF Conference on Computer Vision and Pattern Recognition. pp. 20612--20622 (2024)

\bibitem{bytetrack}
Zhang, Y., Sun, P., Jiang, Y., Yu, D., Weng, F., Yuan, Z., Luo, P., Liu, W., Wang, X.: Bytetrack: Multi-object tracking by associating every detection box. In: Proceedings of the European Conference on Computer Vision (ECCV) (2022)

\bibitem{zhou2022survey}
Zhou, T., Porikli, F., Crandall, D.J., Van~Gool, L., Wang, W.: A survey on deep learning technique for video segmentation. IEEE transactions on pattern analysis and machine intelligence  \textbf{45}(6),  7099--7122 (2022)

\bibitem{zhou2022detecting}
Zhou, X., Girdhar, R., Joulin, A., Kr{\"a}henb{\"u}hl, P., Misra, I.: Detecting twenty-thousand classes using image-level supervision. In: ECCV (2022)

\bibitem{zhou2020tracking}
Zhou, X., Koltun, V., Kr{\"a}henb{\"u}hl, P.: Tracking objects as points. In: European conference on computer vision. pp. 474--490. Springer (2020)

\end{thebibliography}

\appendix

\section{Implementation details}%
\label{sec:impl}

In the EPIC Fields~\cite{EPICFields2023} dataset, the per-frame camera pose and intrinsics are obtained using COLMAP~\cite{colmap}, which also provides a sparse point cloud representing the \textit{static} parts of the scene.
We follow~\cite{plizzari2023}, and obtain the depth maps, $\mathbf{D}_t$, by first calculating a mesh from the sparse point clouds, and then aligning the predictions of a pre-trained depth model to the mesh through shift and scale transformations.
We use Depth Anything~\cite{depthanything}, a state-of-the-art monocular depth estimation model.
We obtain the scale-shift parameters for each depth map by optimizing the L1 distance between the transformed depth map and the rasterized mesh depth.
We use the DINOv2~\cite{oquab2023dinov2} encoder to compute the visual features for segments.
When optimizing the tracking cost, the visual feature for a ``track'' is computed as the average of the visual features of the most recent $100$ observations assigned to the track.

\subsection{Details on frame-by-frame Tracking Cost optimization}%
\label{sec:detail-track}

Our tracking algorithm processes the video sequentially, applying the cost optimization frame-by-frame.
At each frame $t$, we consider:
\begin{enumerate}
    \item $M$ existing tracks from the previous $t-1$ frames
    \item $N$ new observations from the current frame $t$
\end{enumerate}

Here, an observation refers to the set of attributes for a segment (Section 3.2 of main paper), while a track is a sequence of observations that have been assigned to the same instance across frames.
We employ the Hungarian algorithm to perform matching between the $M$ existing tracks and $N$ new observations.
This matching process is guided by our cost formulation (Section 3.3).

For new observations that do not match any existing track (\ie their matching cost exceeds the threshold $\gamma$), we initialize new tracks.
This allows our method to accommodate the introduction of new objects into the scene.

Importantly, we do not terminate tracks that fail to match with a new observation in the current frame.
Instead, we maintain these tracks in our database, propagating their attributes from time $t-1$ to time $t$.
This approach allows our method to handle temporary occlusions or brief disappearances of objects, maintaining object identity over longer periods.

This process enables effective tracking of multiple objects across extended video sequences, addressing challenges like object entries, exits, and occlusions.

\subsection{Hyperparameters}%
\label{sec:hyperparam}

Our model has five hyperparameters: $\gamma$, $\alpha_s$, $\alpha_v$, $\alpha_l$, $\alpha_c$.
We set $\alpha_c=10^4$ and $\gamma=30$ based on observed cost values.
The remaining parameters were tuned on a held-out set of 4 videos, yielding optimal values of $\alpha_s=10,\alpha_v=2,\alpha_l=10$.
These settings were used across all experiments.

\subsection{Evaluation Data}%
\label{sec:eval}

We evaluate our method and baselines using the VISOR dataset, which provides pixel-level annotations for active objects in kitchen environments.
These annotations include any objects used for cooking or cleaning.
From these annotations, we derive ground truth tracks and segmentations.
The dataset's annotation structure supports instance-level tracking, as segments of a particular object category often correspond to the same instance throughout a video.
As mentioned in the main paper, we evaluate our approach using the VISOR annotations for 20 videos from EPIC Fields~\cite{EPICFields2023} dataset.

\section{Additional results on the Ego4D dataset}

To further demonstrate our method's applicability, we also include results on a few select scenes from the Ego4D~\cite{grauman2022ego4d} dataset. We follow the same evaluation protocol used with the EPIC Fields dataset and utilize Egotracks~\cite{tang2024egotracks} for the ground-truth track annotations. 

\Cref{tab:ego4d} shows that our method can consistently refine the tracks obtained by DEVA~\cite{cheng2023tracking} on 3 videos. We use the same hyperparameters for these experiments as the ones used with the EPIC Fields datasets described in \cref{sec:hyperparam}. The video lengths are 2-3$\times$ shorter compared to EPIC Fields, which we believe results in smaller margins of improvement as compared to the ones showed in the main paper.

\begin{table}[!h]
\centering
\caption{Results on the Ego4D~\cite{grauman2022ego4d} dataset.}%
\label{tab:ego4d}
\begin{tabular}{c cccc cccc}
\toprule
\multirow{2}{*}{Video ID} & \multicolumn{4}{c}{DEVA~\cite{cheng2023tracking}} & \multicolumn{4}{c}{Ours (w/ DEVA)} \\
\cmidrule(r){2-5}
\cmidrule(lr){6-9}

& HOTA  & DetA  & AssA  & IDF1  & HOTA  & DetA  & AssA  & IDF1  \\
\midrule
\makecell{\texttt{8b47ac19-7c4f-47d2-}\\\texttt{b5d0-755b524b66b2}}  & 15.29 & 12.26 & 22.74 & 13.56 & \textbf{17.40} & 12.22 & \textbf{28.15} & \textbf{14.98} \\
\makecell{\texttt{9f5253af-acc3-40ca-}\\\texttt{b8bf-7b931f875bd7}} & 12.37 & 9.25 & 18.61 & 10.43 & \textbf{14.38} & 9.25 & \textbf{23.04} & \textbf{12.33} \\
\makecell{\texttt{bff3d583-ca3b-44b8-}\\\texttt{9740-3b34c5a8d7a9}}  & 21.58 & 13.60 & 36.21 & 18.27 & \textbf{23.73} & 13.58 & \textbf{43.96} & \textbf{20.16} \\
\bottomrule
\end{tabular}
\end{table}

\section{On-device Inference Runtime Analysis}

Our method processes egocentric videos in an online manner.
While we compute meshes in advance, we track the objects online with the entire pipeline running at 20FPS on a A6000 GPU\@.
Each of our method's components runs as follows: DINOv2 at 43FPS, the lifting to 3D with DepthAnythingV2 at 23FPS, the prediction of segmentation masks with OWLv2 (run every 5 frames) at 31FPS, the temporal propagation with DEVA~\cite{cheng2023tracking} at 25 FPS\@.
We use \verb!torch.cuda.stream! for asynchronous execution of all models on the same GPU.

\section{Sensitivity to Hyperparameters}
\begin{figure}[t]
    \centering
    \includegraphics[width=\textwidth]{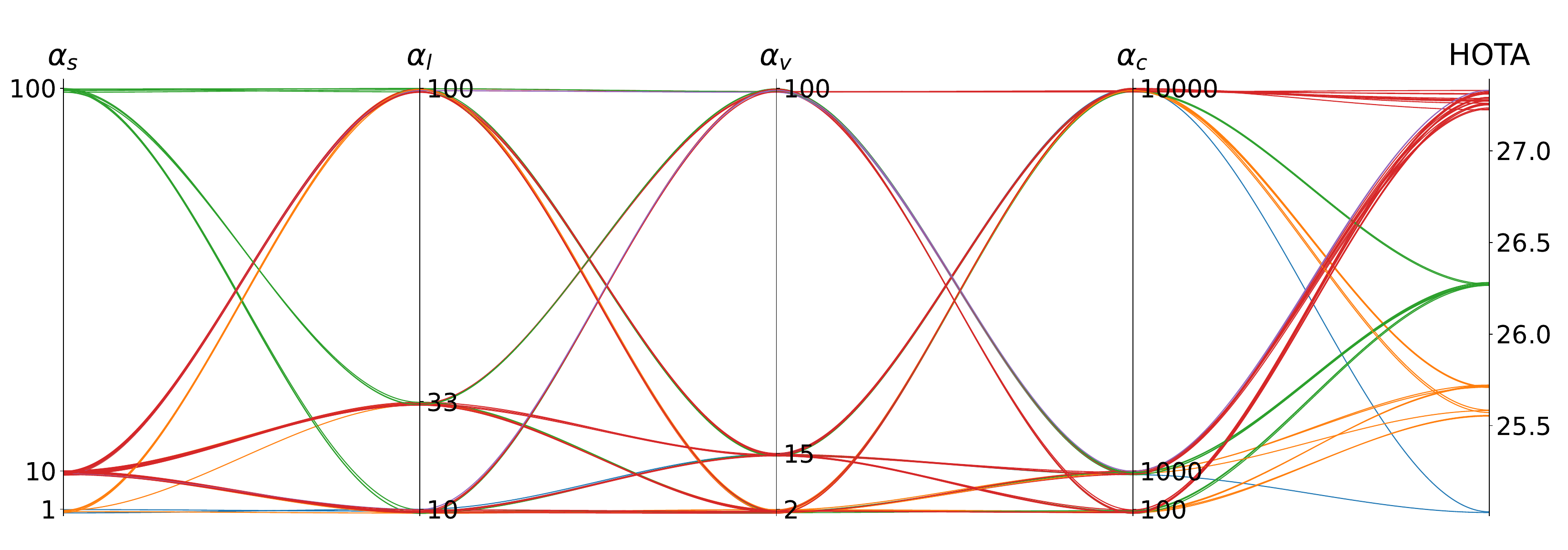}
    \vspace{-1.5em}
    \caption{Sensitivity analysis of HOTA performance to hyperparameters.
    Each vertical axis represents a hyperparameter ($\alpha_s,\alpha_l,\alpha_v,\alpha_c$) or the HOTA metric (\textit{rightmost} axis).
    Colored lines show individual configurations, where intersections with the vertical axes indicating parameter values and resulting HOTA scores.}%
    \label{fig:gridsearch}
\end{figure}

We evaluate the sensitivity of our method by varying the values of the 4 hyperparameters: $\alpha_s$, $\alpha_c$, $\alpha_l$, $\alpha_v$ in the cost-matching formulation.
We perform this analysis on a subset of 5 videos, using 3 representative values for each hyperparameter, resulting in $3^4=81$ configurations.
\Cref{fig:gridsearch} shows that $57$ out of these $81$ hyperparameter configurations lead to a HOTA score in the range $27.2\pm 0.2$, which shows the robustness of our method to these parameters.
There are some configurations, \eg when $\alpha_s=100$ or $\alpha_c=100$, that lead to a degradation in performance.

\section{Ablation without \textit{category} and \textit{instance} terms}
\label{sec:osnom}

Here, we ablate our method by disabling the instance and category terms in the cost formulation, relying solely on 3D location and visual feature costs. Note that this ablation is compared with the ``full'' version of our method on the object tracking task.

\begin{table}[h]
\centering
\caption{Ablation of our method without \textit{category} and \textit{instance} terms in the cost formulation, evaluated on the egocentric object tracking task.}%
\label{tab:osnom-comp}
\begin{tabular}{lcccc}
\toprule
Method & HOTA & DetA & AssA & IDF1 \\
\midrule
DEVA~\cite{cheng2023tracking} & 25.14 & 18.40 & 36.72 & 22.17 \\
Ours (without \textit{cat} and \textit{ins} terms) & 11.31 & 6.49 & 24.16 & 12.98 \\
Ours (full version) & 27.72 & 18.38 & 43.90 & 26.63 \\
\bottomrule
\end{tabular}
\end{table}

As shown in \Cref{tab:osnom-comp}, this ablation results in a significant performance drop, with metrics falling below even the initial performance obtained using the base 2D model, DEVA~\cite{cheng2023tracking}. This is due to two reasons.
First, without the ``instance'' cost, the model completely ignores the initial tracks provided by DEVA\@.
Second, without the ``category'' cost, the model often confuses objects across categories (\eg pot vs sink, knife vs spoon).
Since the Detection Accuracy (DetA) for a video/scene is computed on predicted instances per class, this leads to a severely low DetA on account of various misclassified instances.
This comparison underscores the importance of our additional cost terms in maintaining robust long-term tracking performance in egocentric settings.

Note, the ablation brings the method somewhat closer
to the the OSNOM-like configuration of~\cite{plizzari2023}. 
Although our method was inspired by OSNOM, we can't carry out a direct comparison as the authors of~\cite{plizzari2023} have not yet released their code, or details of the videos used in their evaluation.

\section{Downstream applications}

Our 3D-aware instance segmentation and tracking method yields longer and more consistent tracks than 2D methods. This improvement enables two key downstream applications: 3D object reconstruction and amodal segmentation.

\paragraph{Reconstruction of objects.}

The longer, more consistent tracks produced by our method allow us to extract the same object from many frames using the output instance ID\@.
This multi-view information is crucial for achieving high-quality 3D reconstructions.
This is something that fragmented or inconsistent tracks from 2D methods often fail to achieve.
Additionally, our 3D tracking approach, which uses lifted centroids, allows us to determine the time ranges when an object remains static.
We leverage these static periods for reconstruction, as they provide the most reliable geometric information.
This selective use of frames is only possible due to maintaining long-term tracks of objects.

\paragraph{Amodal segmentation.}

Amodal segmentation aims to estimate the full extent of objects, including parts that are occluded.
Building upon our 3D object reconstructions, we render the reconstructed 3D object from multiple viewpoints corresponding to different frames in the video.
This process allows us to generate occlusion-free, amodal segmentations of the object.

\begin{figure}[t]
    \centering
    \includegraphics[width=\textwidth]{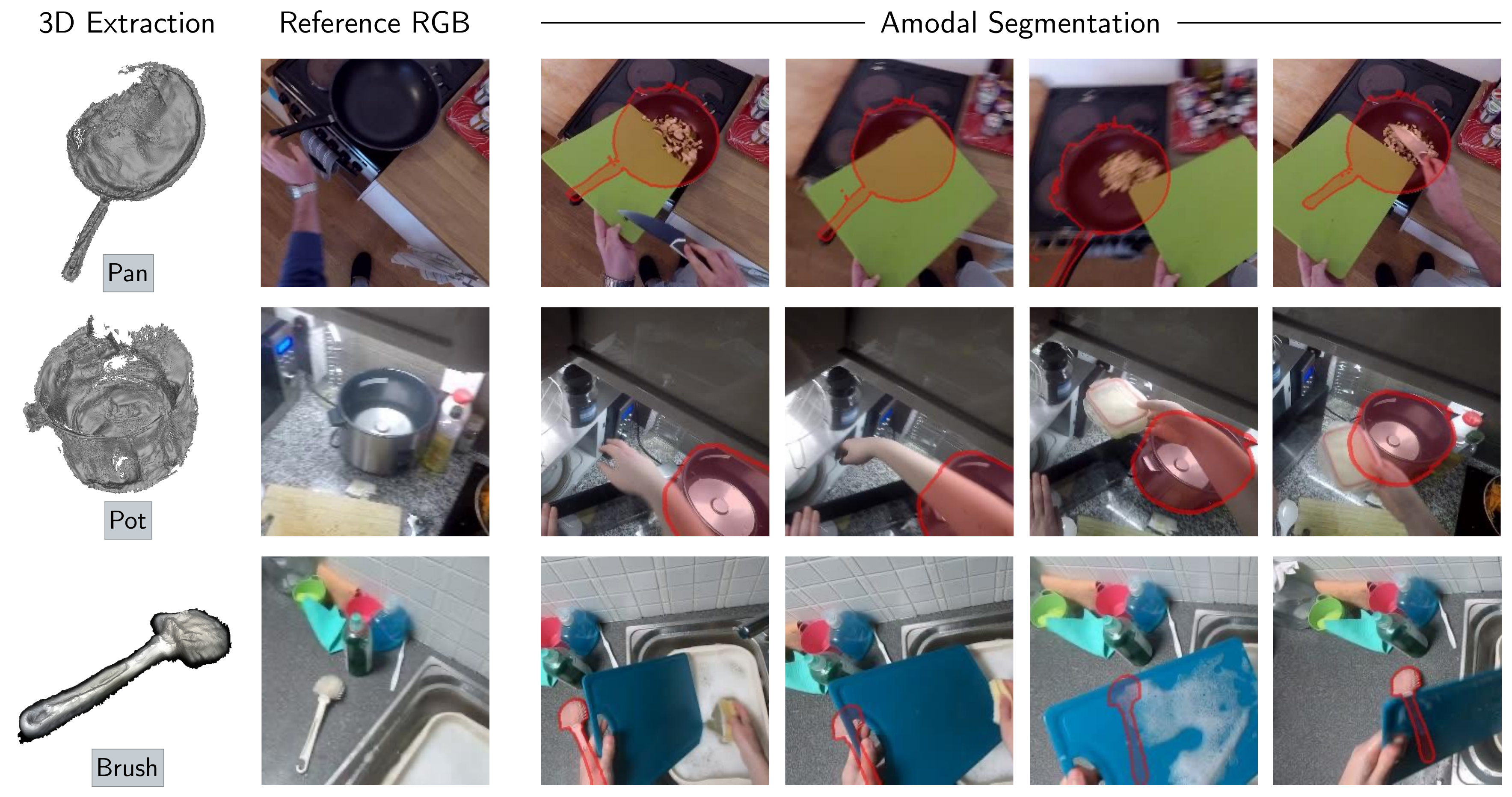}
    \caption{Qualitative results demonstrating the quality of object reconstructions and amodal segmentations obtained using our 3D-aware tracking method.
    The ``Reference RGB'' column show an image containing the referred object \textit{unoccluded}.
    Last 4 columns show the resulting amodal segmentations of the object in \textcolor{red}{red} masks with a \textcolor{red}{\textbf{red}} border.}%
    \label{fig:amodal_seg}
\end{figure}

These applications demonstrate the cascading benefits of our improved 3D-aware tracking method.
We show qualitative results in \cref{fig:amodal_seg} that demonstrate the quality of object reconstructions and amodal segmentations obtained using our method.
In practice, we use the 2D Gaussian Splatting~\cite{Huang2DGS2024} approach to obtain precise mesh reconstructions for these objects.

\section{Details on Obtaining Amodal Segmentations}%
\label{sec:amodal}

This section elaborates on the process of obtaining amodal segmentations, which involves three main steps: identifying static object frames, 3D object reconstruction, and amodal segmentation projection.

\paragraph{\textbf{Identifying Static Object Frames}:}

We begin by analyzing the tracked 3D centroid of the object of interest across the video sequence.
By identifying periods where the centroid remains relatively stationary (using a threshold on 3D location differences between frames), we can isolate a range of frames where the object is static.
This step is crucial as it allows us to gather multiple views of the object from different camera angles while minimizing the complexity introduced by object motion.

\paragraph{\textbf{3D Object Reconstruction}:}

Once we have identified the static frames, we utilize the corresponding 2D instance segmentations and associated camera parameters to reconstruct the 3D shape of the object.
This reconstruction is achieved through a technique known as Gaussian Splatting\footnote{We use 2D Gaussian Splatting~\cite{Huang2DGS2024} which is a variation of 3D Gaussian Splatting~\cite{kerbl3Dgaussians} that makes it more straightforward to obtain object meshes.}.
In this approach, we represent the 3D object 
as a collection of Gaussian functions in 3D space.
Each Gaussian is characterized by its mean position and covariance matrix, which define its location and shape respectively.
Given $G$ as the set of 3D Gaussians and a camera viewpoint $C_i$, the differentiable Gaussian Splatting renderer~\cite{kerbl3Dgaussians} produces an image 
\[\hat{I}=\Pi(G,C_i)\in\mathbb{R}^{H \times W \times 3}\]
The same renderer can be used to render an alpha-map (equivalent to a segmentation map) by setting the colors for each Gaussian to be $1$.
The Gaussian Splatting model for the object of interest is optimized by minimizing this loss function across multiple views:
\[L = \sum_t (I_t \odot m_t - \Pi(G, C_i))^2\]
where $m_t$ and $I_t$ represents the observed 2D segmentation map and RGB values in frame $t$ respectively.

\paragraph{\textbf{Projecting Amodal Segmentations}:}

Once we obtain a satisfactory 3D reconstruction of the object, we can generate amodal segmentations for any desired viewpoint.
This is done by rendering the entire 3D Gaussian representation back onto the image plane, regardless of occlusions present in the original views.
As explained above, we set the Gaussian \textit{colors} to $1$ which provides an alpha map using the renderer as 
\[
\hat{m}=\Pi(G,C_i)\in\mathbb{R}^{H \times W}
\]
where $\hat{m}$ is the amodal segmentation map.
This map represents the full extent of the object, including parts that may be occluded in the original views.
The values in $\hat{m}$ range from 0 to 1, indicating the likelihood of each pixel belonging to the object.

This approach allows us to generate accurate amodal segmentations that account for the full 3D structure of the object, providing a more complete representation than what is directly observable in any single frame of the video.

\section{Limitations}

Our method significantly improves object tracking in egocentric videos, especially under conditions of rapid motion, occlusions, and out-of-sight objects. However, there are limitations, particularly in scenarios where accurate camera poses are difficult to obtain or estimate. Specifically, our approach relies heavily on the assumption that high-quality camera intrinsics and extrinsics are available, as they are essential for accurate 3D lifting of object segments. Hence, performance can degrade in cases with noisy depth maps or challenging conditions like motion blur, poor lighting, or extreme viewpoint changes, as these factors reduce the precision of 3D reconstruction.
\end{document}